\documentclass{article}

\PassOptionsToPackage{numbers, compress}{natbib}

\usepackage[final]{nips_2018}

\usepackage[utf8]{inputenc} 
\usepackage[T1]{fontenc}    
\usepackage{hyperref}       
\usepackage{url}            
\usepackage{booktabs}       
\usepackage{amsfonts}       
\usepackage{nicefrac}       
\usepackage{microtype}      
\usepackage{graphicx}
\usepackage{subfigure}
\usepackage{amsmath}
\usepackage[table,xcdraw]{xcolor}
\definecolor{darkblue}{rgb}{0,0.08,0.45}
\hypersetup{
    colorlinks=true,
    citecolor=darkblue
}

\title{Perceiving Physical Equation by Observing Visual Scenarios}

\author{
  Siyu Huang\thanks{Equal contributions. This work was done when Siyu Huang and Zhi-Qi Cheng were visiting Carnegie Mellon University.} \\
  Zhejiang University\\
  \texttt{siyuhuang@zju.edu.cn}\\ 
  \And
  Zhi-Qi Cheng\footnotemark[1]\\
  Southwest Jiaotong University\\
  \texttt{zhiqicheng@gmail.com}\\ 
  \And
  Xi Li\\
  Zhejiang University\\
  \texttt{xilizju@zju.edu.cn}\\ 
  \And
  Xiao Wu\\
  Southwest Jiaotong University\\
  \texttt{wuxiaohk@gmail.com}\\ 
  \And
  Zhongfei (Mark) Zhang\\
  Zhejiang University\\
  \texttt{zhongfei@zju.edu.cn}\\ 
  \And
  Alexander Hauptmann\\
  Carnegie Mellon University\\
  \texttt{alex@cs.cmu.edu}\\ 
}

\begin{document}

\maketitle

\begin{abstract}
Inferring universal laws of the environment is an important ability of human intelligence as well as a symbol of general AI. In this paper, we take a step toward this goal such that we introduce a new challenging problem of inferring invariant physical equation from visual scenarios. For instance, teaching a machine to automatically derive the gravitational acceleration formula by watching a free-falling object. To tackle this challenge, we present a novel pipeline comprised of an Observer Engine and a Physicist Engine by respectively imitating the actions of an observer and a physicist in the real world. Generally, the Observer Engine watches the visual scenarios and then extracting the physical properties of objects. The Physicist Engine analyses these data and then summarizing the inherent laws of object dynamics. Specifically, the learned laws are expressed by mathematical equations such that they are more interpretable than the results given by common probabilistic models. Experiments on synthetic videos have shown that our pipeline is able to discover physical equations on various physical worlds with different visual appearances.

\end{abstract}

\section{Introduction}
Inference is one of the most basic and significant aspects of human intelligence \cite{tenenbaum2011grow} as well as AI \cite{lake2017building}. As a high-level aspect of inference, the induction of universal laws from observations of our world is both the core basis and the goal of the scientific research. For example, Sir Isaac Newton saw an apple falling down and then was inspired to discover the law of gravitation. However, for a computing machine, the induction of laws based on visual observations is still a very challenging and open problem, and has been rarely explored by the existing literature until today. 

In this paper, we introduce a new problem that we attempt to teach machine to automatically derive mathematical expressions of object dynamics from videos of a physical world. In contrast to the most recent approaches \cite{battaglia2016interaction,watters2017visual,wu2017learning} which explores to learn object mechanical behaviors by the black box of deep neural networks, we aim at explicitly presenting the symbolic expressions of latent physical laws, leading to a more interpretable model and more visualizable results. A pioneer work \cite{schmidt2009distilling} learns to derive mathematical equations from the data of physical experiments. While in this work, we propose to learn mathematical expressions directly from complicated videos.   

Toward this goal, we propose a novel pipeline comprised of an \emph{Observer Engine} and a \emph{Physicist Engine}. The Observer Engine acts like an observer that watches the videos of a physical scenario and extracts the physical properties of objects in that scenario. Then the Physicist Engine imitates a physicist that summarizes the observed data and finally derives the mathematical equations. 

In the experiments, we evaluate our pipeline on synthetic videos of multiple physical scenarios, showing that it is able to learn precise mathematical equations on these physical worlds with diverse visual appearances. We also explore several variants of models for the Observer Engine and the Physicist Engine respectively, so as to quantitatively establish baselines for relevant research in the future.

Our contributions are three-fold. First, we introduce a new problem of learning mathematical equations of object dynamics from videos, taking a step toward the automatic induction of universal laws for general AI. Second, we propose a novel pipeline to tackle this challenging problem. Third, empirical studies demonstrate the effectiveness of our approach on several synthetic physical scenarios.


\section{Related Work}
\paragraph{Physical Reasoning}
Physical reasoning has drawn much attention of AI researchers in recent years. Previous work on physical reasoning explored to learn the common sense knowledge of physical scenarios \cite{battaglia2013simulation,goyal2017something} and to develop the simulation techniques for inferring the future states of physical systems \cite{fragkiadaki2015learning,mottaghi2016happens,wang2018learning}. A typical example is to predict whether a stack of blocks would fall \cite{gupta2010blocks,lerer2016learning,li2016fall}. In addition, the simulation and prediction of macroscopic physical phenomena, including weather events \cite{racah2017extremeweather} and fluid \cite{jeong2015data,de2017deep}, were also studied by researchers. The ``NeuroAnimator'' \cite{grzeszczuk1998neuroanimator} was the pioneer work to quantitatively simulate the physical dynamics of articulated bodies with neural networks. Today, the learning of object dynamics \cite{mottaghi2016newtonian,byravan2017se3,chang2016compositional,ehrhardt2017learning} becomes a research hotspot. 

More recently, researchers incorporated the powerful deep neural networks into physical reasoning systems to enable a deeper understanding of the physical properties underlied in visual scenarios. Interaction Network (IN) \cite{battaglia2016interaction} and Visual Interaction Network (VIN) \cite{watters2017visual} were successively proposed for modeling the dynamic relationships between physical objects in videos. Wu et al. \cite{wu2017learning} end-to-end learned a hybrid of graphics engines and physics engines to predict the long-term visual observations of a physical world. 

\begin{figure}[t]
\centering
\includegraphics[width=0.95\linewidth]{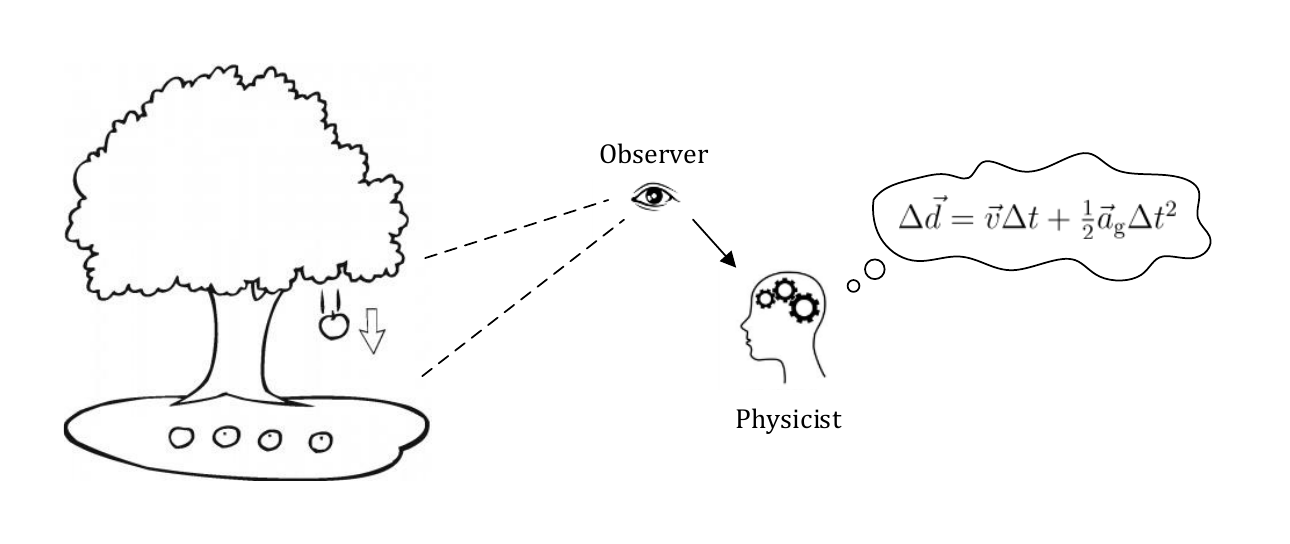}
\caption{\textbf{Observing and thinking:} inferring physical equation from visual scenario. The ability to infer universal law of the environment is one of the significant high-level aspects of human intelligence.
}
\label{fig1}
\end{figure}

In these approaches, the dynamics of objects and their interactions are generally modeled by the non-linear transformations of neural networks which are black-box models. The explicit symbolic expressions of object kinetic properties are not revealed and interpreted. In this work, we take the first step toward the interpretable physical reasoning model in which we attempt to summarize the object kinetic properties as precise physical equations through observing videos of a physical world.

\paragraph{Equation Regression}
In this work, we concentrate on inferring a physical equation from the visual scenarios, which is rarely explored in the existing literature. From the perspective of equation regression, there have been many efforts on learning the symbolic relationships from non-structured data \cite{teodorescu2008high,sutskever2009using,uy2011semantically}. In another aspect, several approaches \cite{bhat2002computing,brubaker2009estimating} learned to fit the parameters of Newtonian mechanics equations to physical systems depicted by videos. For instance, Wu et al. \cite{wu2015galileo,wu2016physics} proposed a deep learning model to infer the physical properties (such as mass, volume, and coefficient friction) of objects from real-world videos. However, the symbolic expression of physical equations themselves are still not learned in these approaches. 

A popular method for the learning of mathematical expression is called ``symbolic regression'' \cite{augusto2000symbolic,giustolisi2006symbolic}, which is adopted in this work for the generation of physical equations. Symbolic regression is a machine learning technique that identifies a mathematical expression to minimize the customized error metric based on genetic programming \cite{koza1994genetic} and evolutionary algorithm \cite{mckay1995using}. Unlike the traditional linear and non-linear regression methods that fit parameters to an equation of a given form, symbolic regression searches both the parameters and the form of equations simultaneously \cite{schmidt2009distilling}. More details of our approach are discussed in the following sections. 

\section{Model}
Our model learns to infer the inherent mathematical equation from video frames of a physical system. It consists of an \emph{Observer Engine} and a \emph{Physicist Engine}.

\begin{figure}[!htb]
\centering
\subfigure[Observer Engine]{
\begin{minipage}[b]{0.63\linewidth}
\includegraphics[width=1\linewidth]{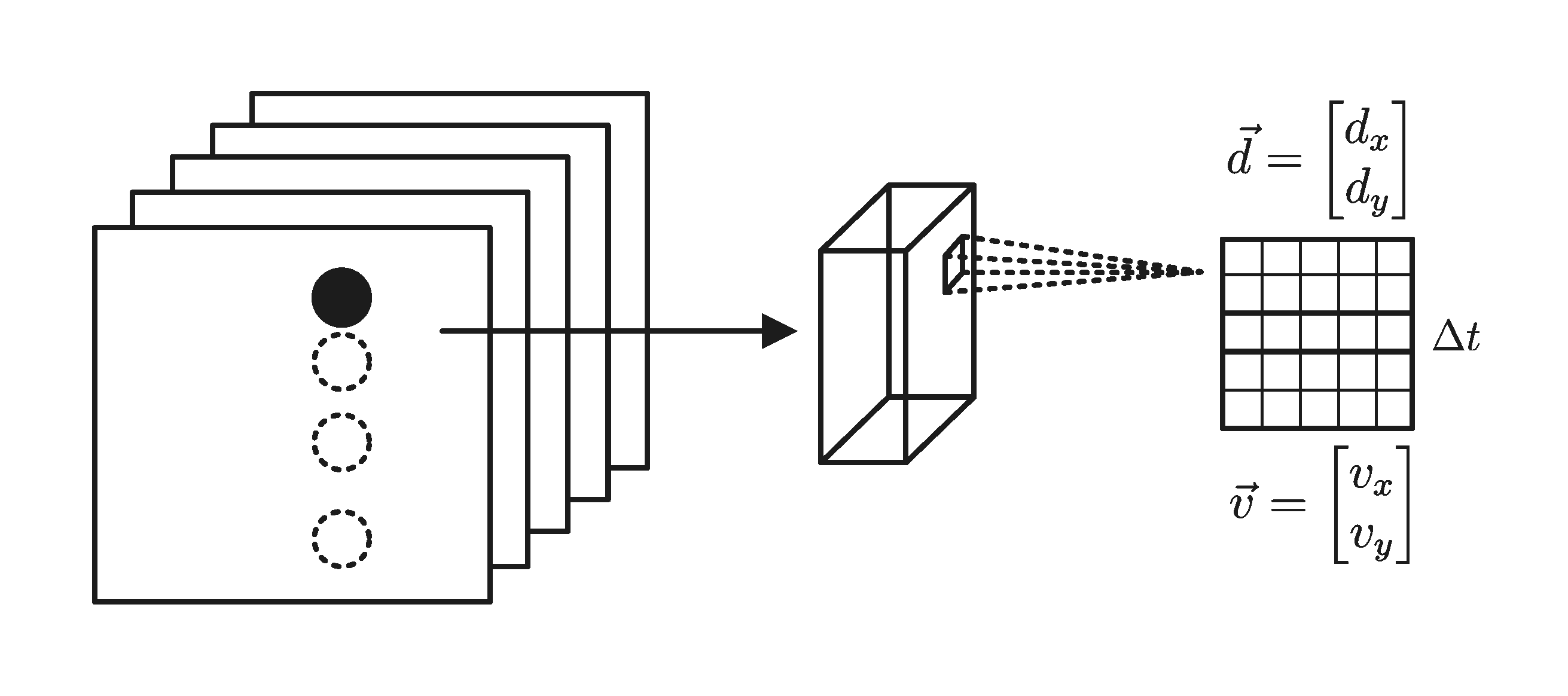}
\end{minipage}
}
\subfigure[Physicist Engine]{
\begin{minipage}[b]{0.3\linewidth}
\includegraphics[width=1\linewidth]{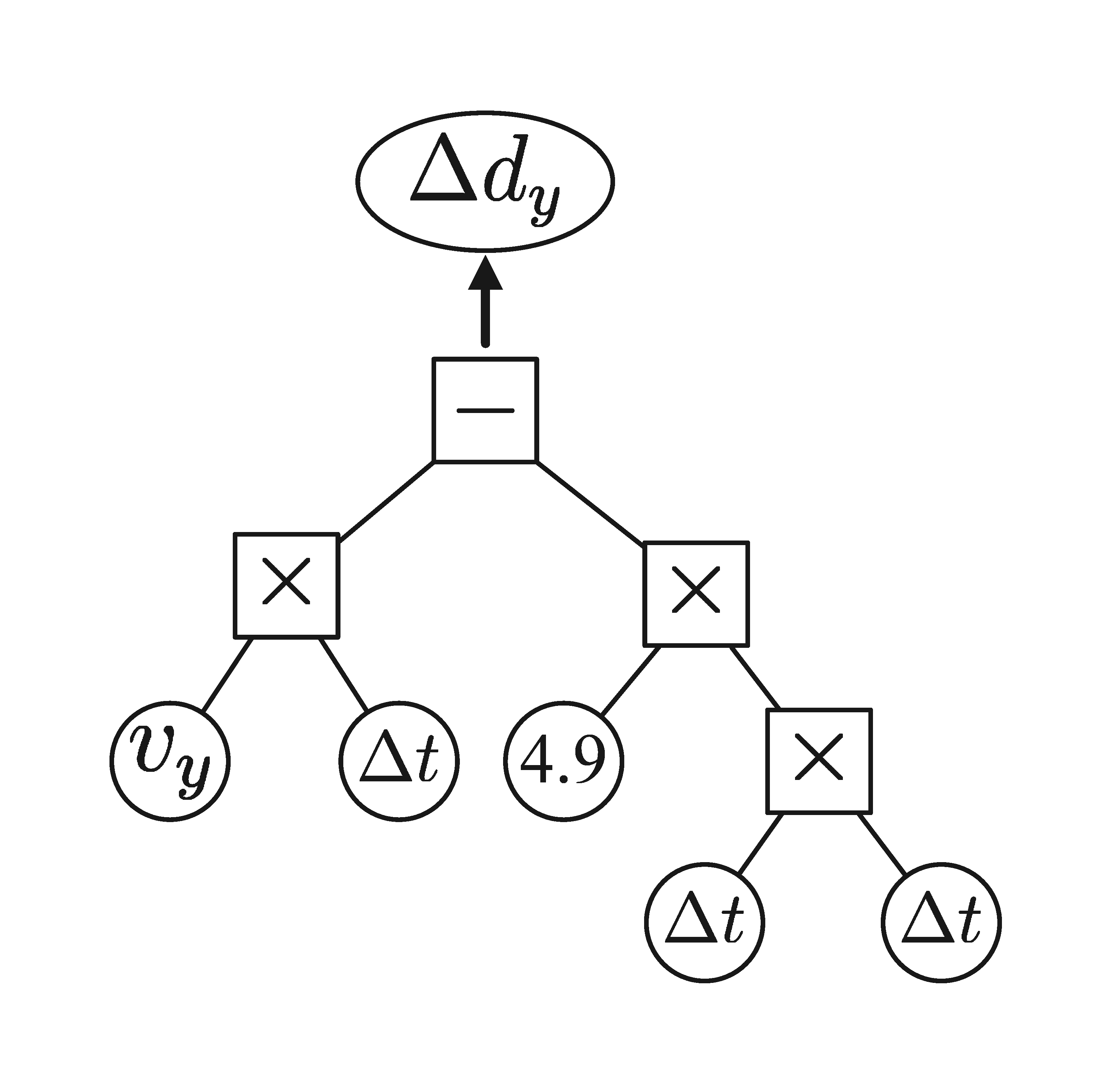}
\end{minipage}
}
\caption{Our model is comprised of (a) the Observer Engine and (b) the Physicist Engine. At left, a video depicts that an object is in free-falling. The Observer Engine uses deep neural networks to extract the physical properties of the object. The Physicist Engine learns a mathematical expression of the object dynamics by evolving a syntax tree based on the property variables.
}
\label{model}
\end{figure}

\paragraph{Observer Engine}
The Observer Engine acts like an observer that watches the videos of a physical world, and at the same time records the physical-property variables. As illustrated in Fig. \ref{model}(a), it captures the physical properties of the kinetic objects and the environment in videos. In this work, we use the Faster-RCNN \cite{ren2015faster} model to detect an object and localize its position $\vec d$ according to coordinates of the bounding-boxes. In order to get a more precise object position, we employ a two-stage approach to refine the position on coarse-to-fine spatial scales. Specifically, a Faster-RCNN detector is applied on an image to get a coarse window of an object, then another Faster-RCNN detector is applied on the window to get a fine bounding-box. The two-stage approach ensures a precise object localization and a speed up of the detection procedure. The velocity $\vec v$ of an object is computed by $\vec v=\Delta \vec d / \Delta t$, where $\Delta t$ is the time interval between two video frames. Observation data $\vec d$, $\vec v$, and $\Delta t$ are fed to the Physicist Engine serving as the independent variables.

\paragraph{Physicist Engine}
The Physicist Engine acts like a physicist that infers the equation based on the observations given by the Observer Engine. It takes a set of objects' physical properties (output from the visual engine applied to a series of videos) as input. It outputs the equation between displacement $\Delta \vec d$ and the independent variables. In this work, we adopt symbolic regression with genetic programming (GP) \cite{augusto2000symbolic,giustolisi2006symbolic} for the inference of mathematical equation, implemented based on GPlearn Toolkit\footnote{http://gplearn.readthedocs.io/en/stable/index.html}. 

As illustrated in Fig. \ref{model}(b), the formula is represented as a syntax tree. The variables, denoted as the round nodes, are leaves of the tree. The mathematical operations, denoted as the square nodes, connect the independent variables. Our goal is to find the best formula consisting of arbitrary independent variables and mathematical operations to minimize the mean absolute error (MAE) corresponding to the given target. At the very beginning, a population of formulas is randomly initialized. In an evolutionary manner, GP evolves the fittest ones of every generation until convergence. More details of GP are discussed in Section \ref{learning_detail}.

\begin{figure}

\centering

\begin{tabular}{c}
\toprule[1pt]\midrule

\subfigure{
\begin{minipage}[c]{0.4\linewidth}
\centerline{\textbf{Physical Scenario}}
\end{minipage}
\begin{minipage}[c]{0.28\linewidth}
\centerline{\textbf{Learned} $\Delta d_x$}
\end{minipage}
\begin{minipage}[c]{0.28\linewidth}
\centerline{\textbf{Learned} $\Delta d_y$}
\end{minipage}
}

\\\midrule

\subfigure{
\begin{minipage}[c]{0.4\linewidth}
\centering
\includegraphics[width=0.32\linewidth]{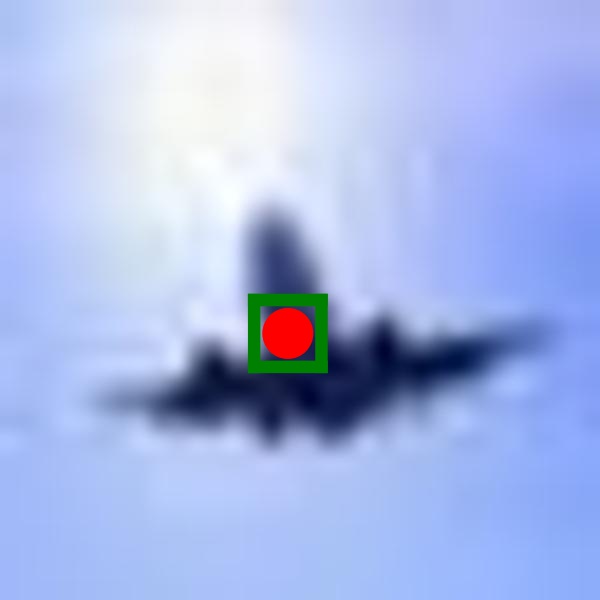}
\includegraphics[width=0.32\linewidth]{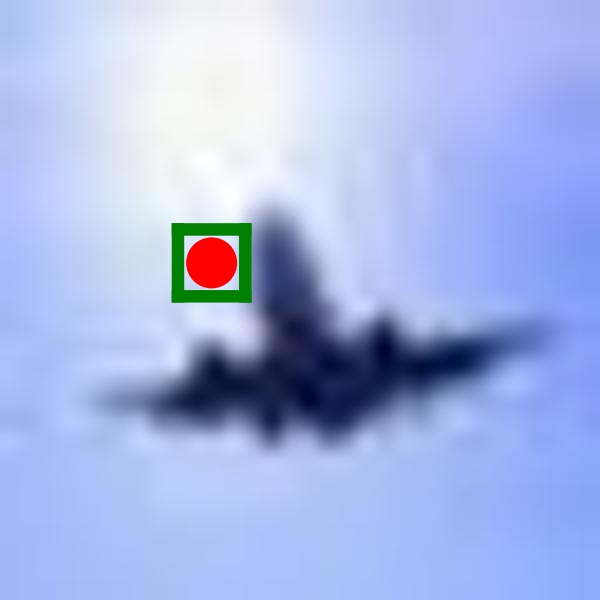}
\includegraphics[width=0.32\linewidth]{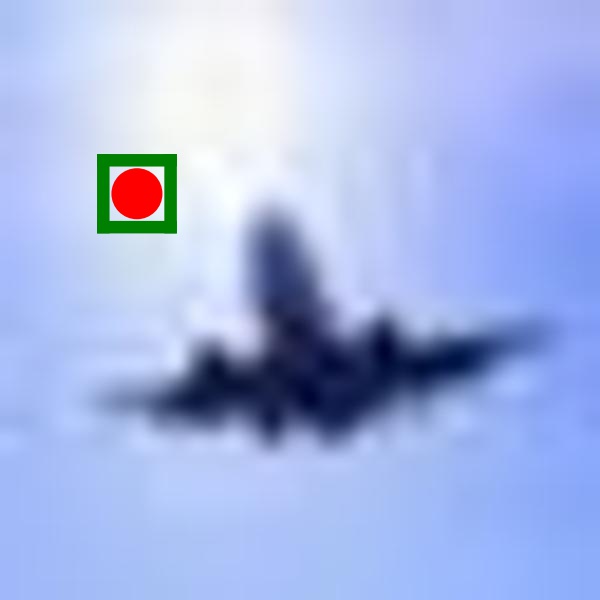}
\centerline{\textbf{Dirft}~(\url{http://bit.ly/2L6JR8W})}
\end{minipage}
\begin{minipage}[c]{0.28\linewidth}
\centering
\includegraphics[scale=0.6]{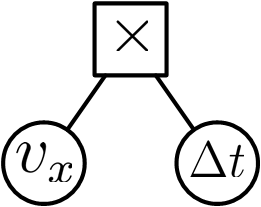}
\small
\begin{equation}
v_x \Delta t \notag
\end{equation}
\end{minipage}
\begin{minipage}[c]{0.28\linewidth}
\centering
\includegraphics[scale=0.6]{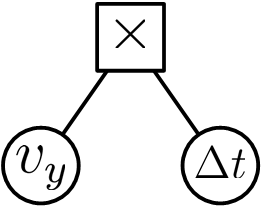}
\small
\begin{equation}
v_y \Delta t  \notag
\end{equation}
\end{minipage}
}

\\\midrule

\subfigure{
\begin{minipage}[c]{0.4\linewidth}
\centering
\includegraphics[width=0.32\linewidth]{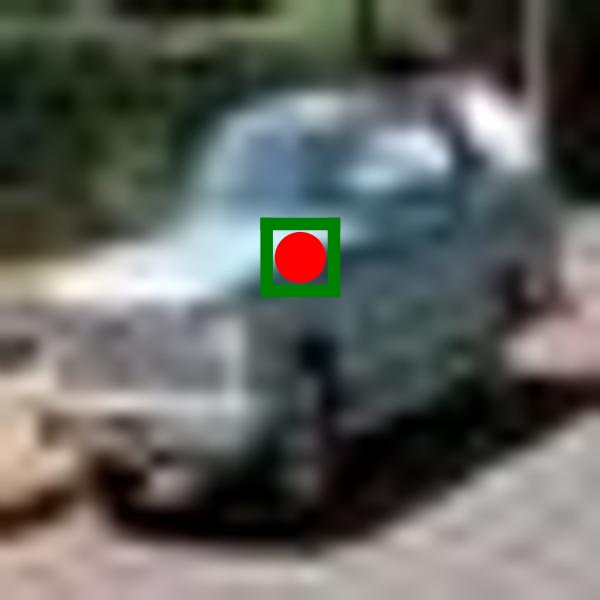}
\includegraphics[width=0.32\linewidth]{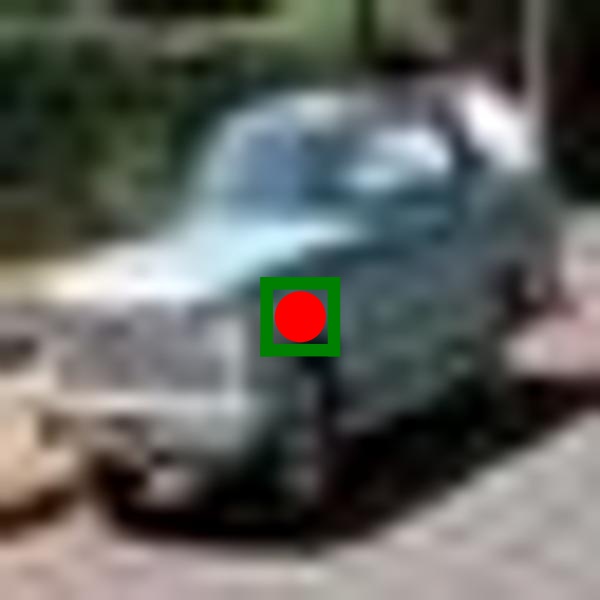}
\includegraphics[width=0.32\linewidth]{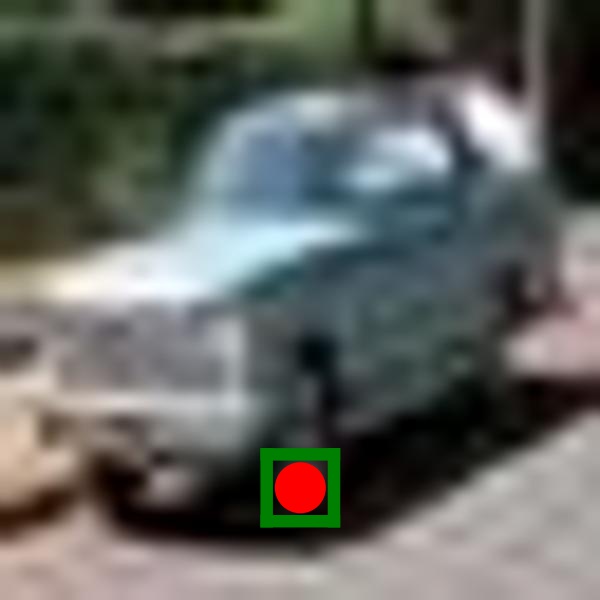}
\centerline{\textbf{Free-falling}~(\url{http://bit.ly/2k3HIyu})}
\end{minipage}
\begin{minipage}[c]{0.28\linewidth}
\centering
\includegraphics[scale=0.7]{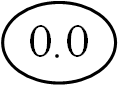}
\small
\begin{equation}
0 \notag
\end{equation}
\end{minipage}
\begin{minipage}[c]{0.28\linewidth}
\centering
\includegraphics[scale=0.55]{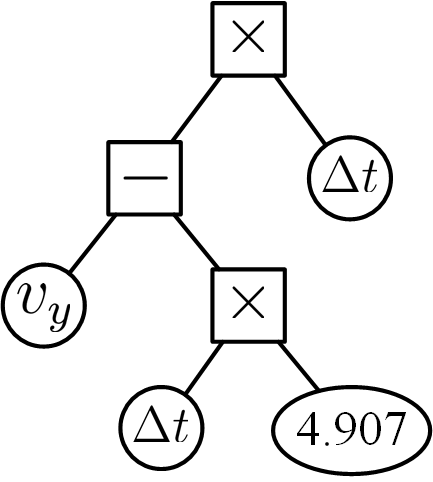}
\small
\begin{equation}
v_y \Delta t - 4.907\Delta t^2 \notag
\end{equation}
\end{minipage}
}

\\\midrule

\begin{minipage}[c]{0.4\linewidth}
\centering
\includegraphics[width=0.32\linewidth]{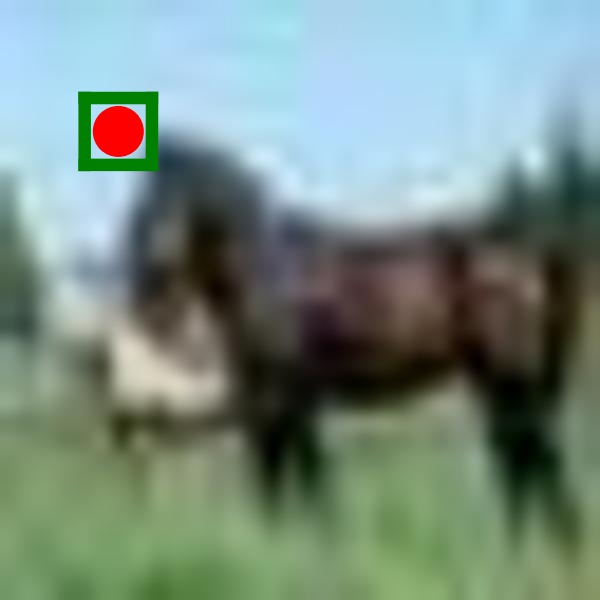}
\includegraphics[width=0.32\linewidth]{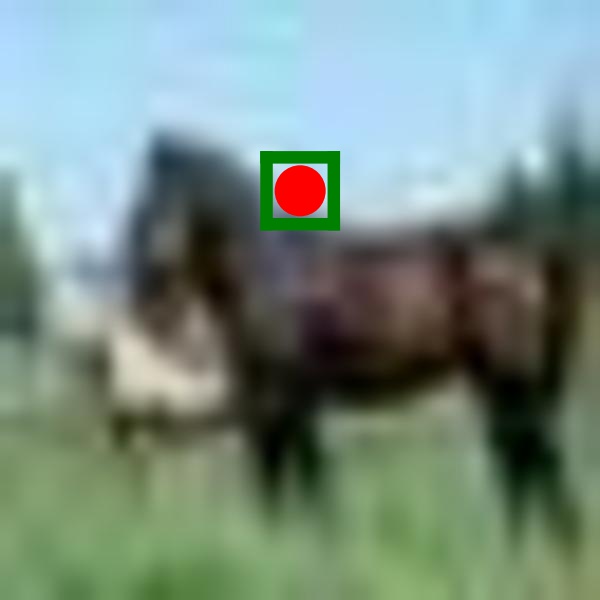}
\includegraphics[width=0.32\linewidth]{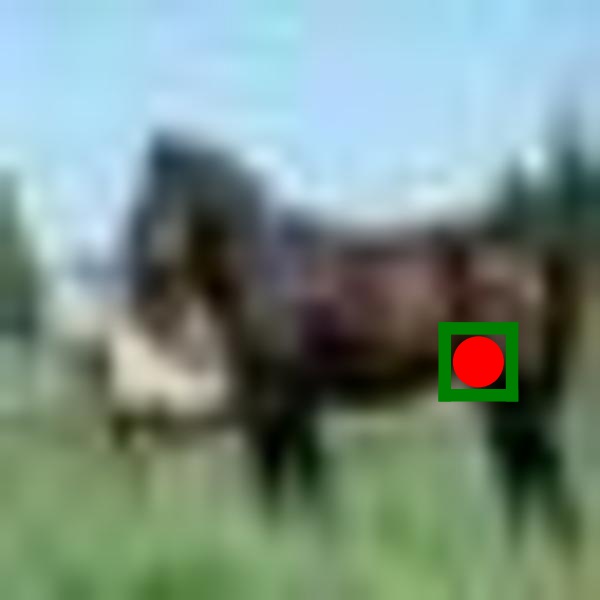}
\centerline{\textbf{Parabola}~(\url{http://bit.ly/2It4Kcq})}
\end{minipage}
\begin{minipage}[c]{0.28\linewidth}
\centering
\includegraphics[scale=0.6]{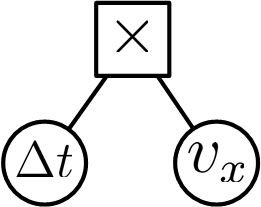}
\small
\begin{equation}
v_x \Delta t \notag \notag
\end{equation}
\end{minipage}
\begin{minipage}[c]{0.28\linewidth}
\centering
\includegraphics[scale=0.55]{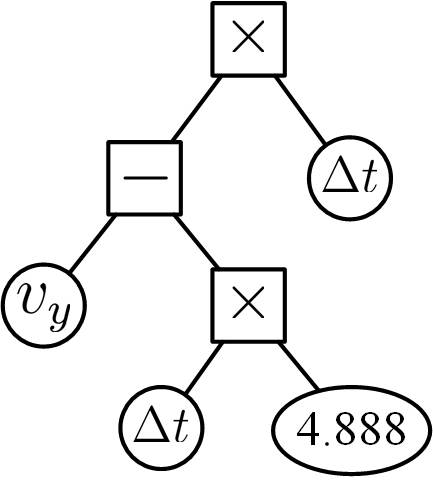}
\small
\begin{equation}
v_y \Delta t - 4.888\Delta t^2 \notag
\end{equation}
\end{minipage}

\\\midrule

\subfigure{
\begin{minipage}[c]{0.4\linewidth}
\centering
\includegraphics[width=0.32\linewidth]{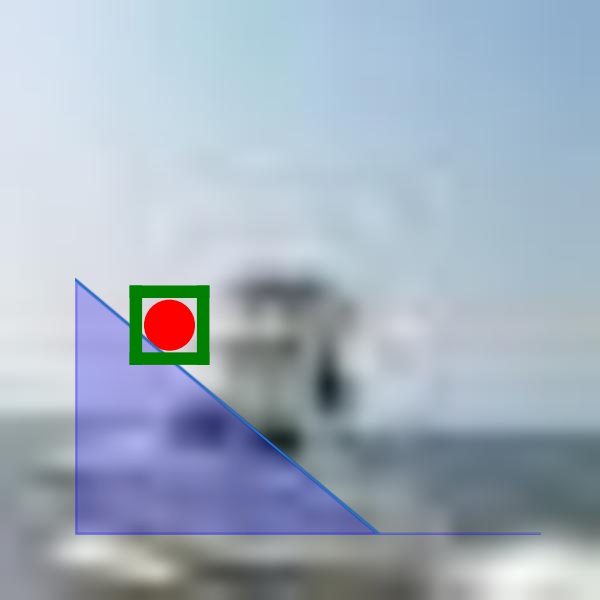}
\includegraphics[width=0.32\linewidth]{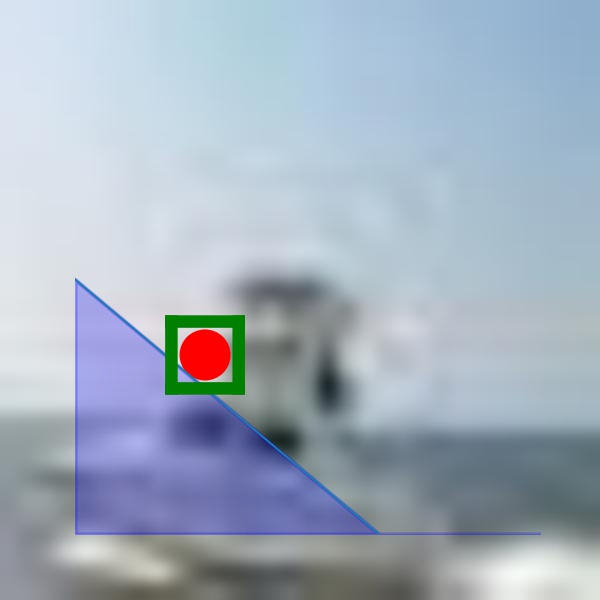}
\includegraphics[width=0.32\linewidth]{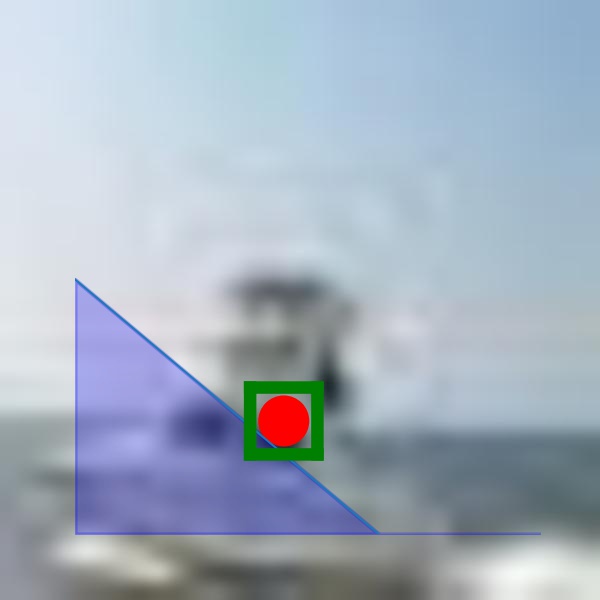}
\centerline{\textbf{Slope}~(\url{http://bit.ly/2L7CIF1})}
\end{minipage}
\begin{minipage}[c]{0.28\linewidth}
\centering
\includegraphics[scale=0.55]{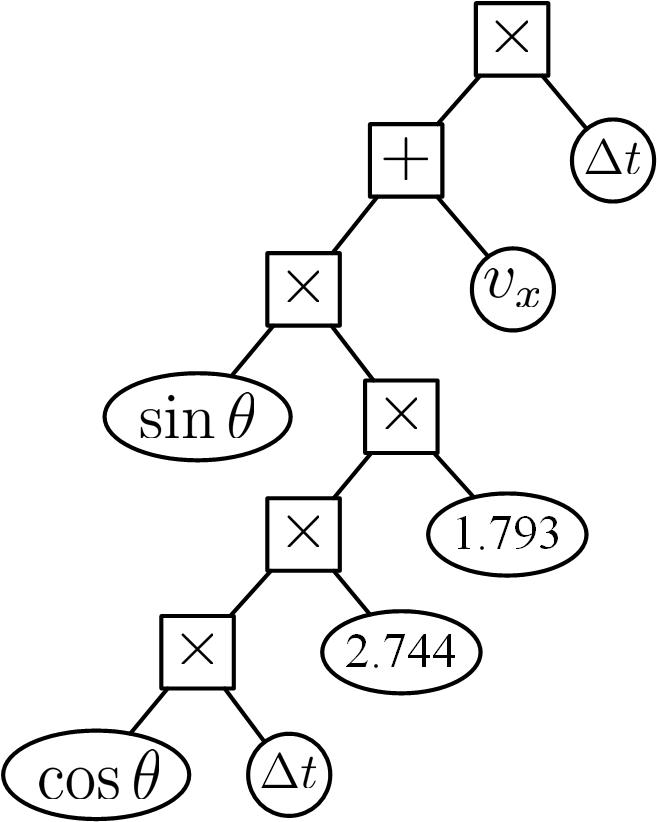}
\small
\begin{equation}
v_x \Delta t + 4.920\sin \theta \cos \theta \Delta t^2 \notag
\end{equation}
\end{minipage}
\begin{minipage}[c]{0.28\linewidth}
\centering
\includegraphics[scale=0.55]{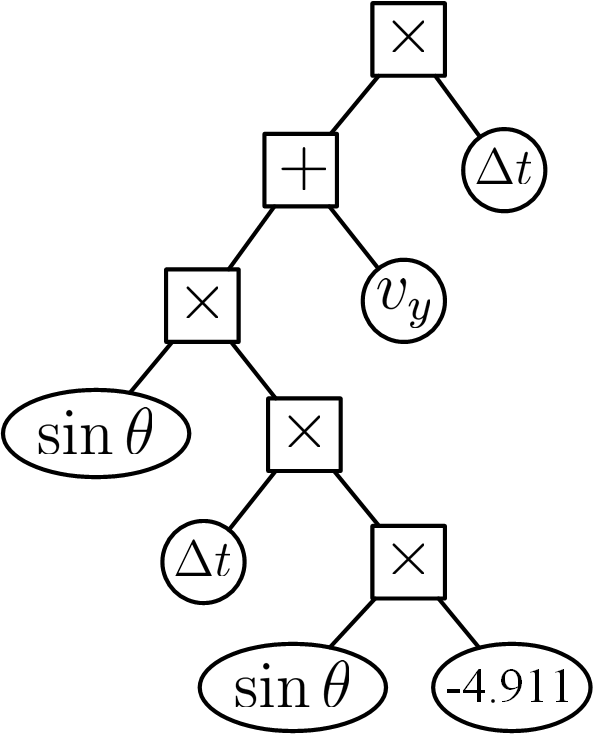}
\small
\begin{equation}
v_y \Delta t - 4.911\sin ^2 \theta \Delta t^2 \notag
\end{equation}
\end{minipage}
}

\\\midrule

\subfigure{
\begin{minipage}[c]{0.4\linewidth}
\centering
\includegraphics[width=0.32\linewidth]{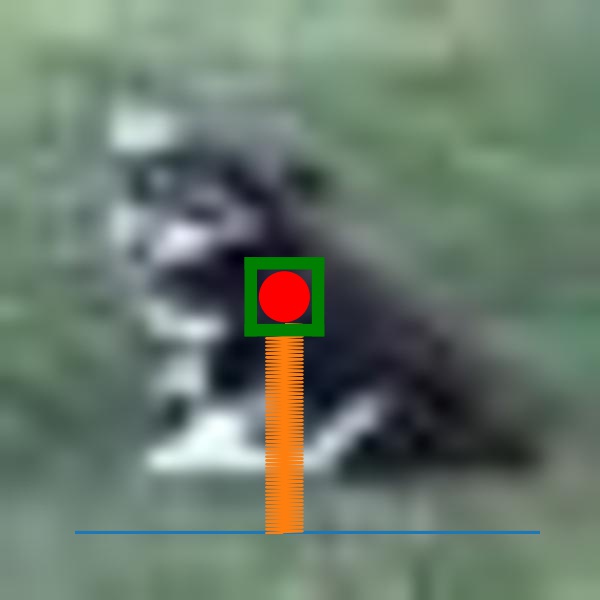}
\includegraphics[width=0.32\linewidth]{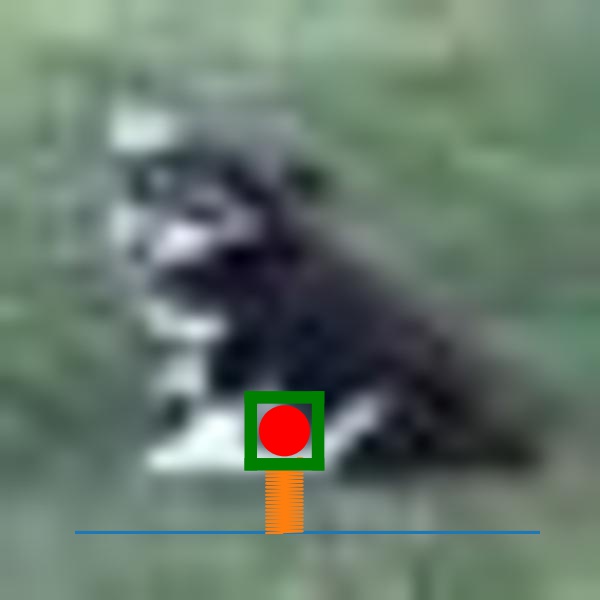}
\includegraphics[width=0.32\linewidth]{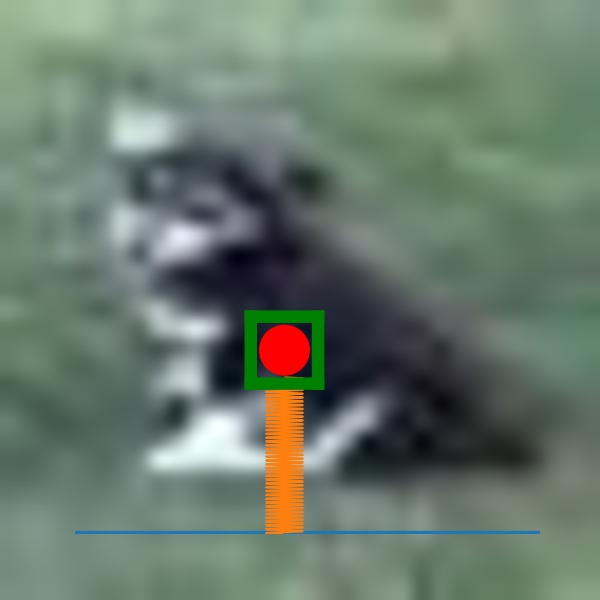}
\centerline{\textbf{Spring}~(\url{http://bit.ly/2KuAfnj})}
\end{minipage}
\begin{minipage}[c]{0.28\linewidth}
\centering
\includegraphics[scale=0.7]{tree_0.png}
\small
\begin{equation}
0 \notag
\end{equation}
\end{minipage}
\begin{minipage}[c]{0.28\linewidth}
\centering
\includegraphics[scale=0.5]{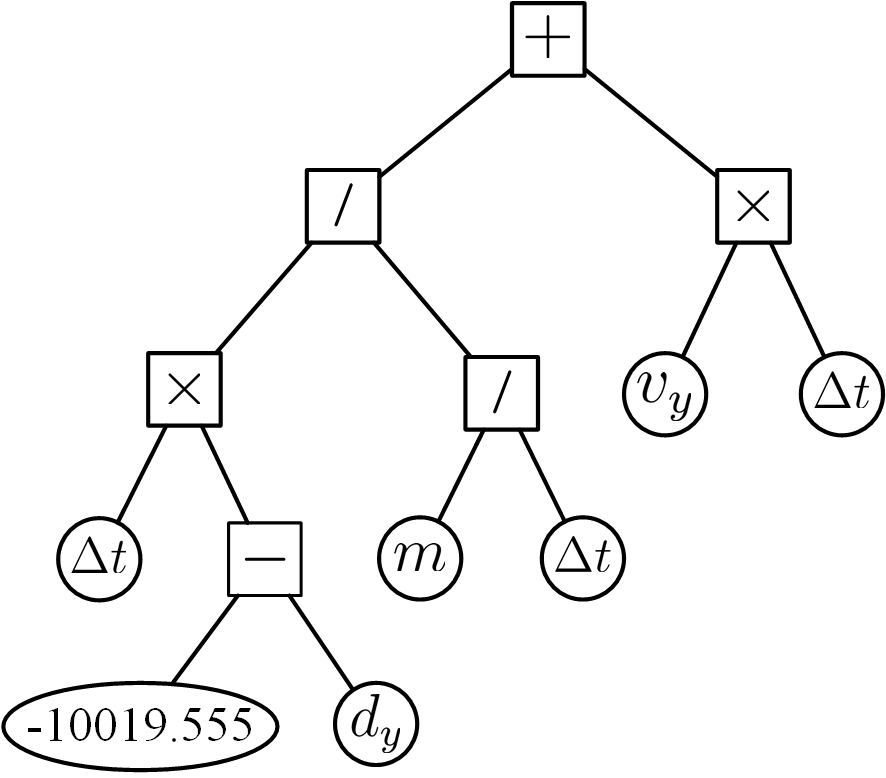}
\small
\begin{equation}
v_y \Delta t - (d_y+10019.555) /m \cdot \Delta t^2 \notag
\end{equation}
\end{minipage}
}

\\ \midrule \bottomrule[1pt]
\end{tabular}

\caption{\textbf{Physical scenarios and our learned equations.} In each scenario, the object moves under particular dynamic equations. Please click the URLs to watch the synthetic videos if interested. Results show that our method is able to learn correct mathematical equations with relatively accurate physical constants in all of the scenarios. The syntax trees are shown together with the equations. The images are resized to 600$\times$600 for visual clarity.}
\label{qualitative_fig}
\end{figure}

\section{Experiments}
\subsection{Physical Scenarios}

We conduct experiments on five types of physical scenarios. In each scenario, there is an object obeying the basic dynamic equation as 
\begin{equation}
\Delta \vec d=\vec v\Delta t+\frac{1}{2}\vec a \Delta t^2
\end{equation}
$\Delta \vec d$ is the displacement vector and $\vec v$ is the velocity vector. $\vec a$ is the accelerated velocity vector corresponding to the specific object dynamics of each physical scenario, including

\begin{itemize}
\item
\textbf{Drift} There is no external force applied on the object. The object drifts with its initial velocity. The accelerated velocity $\vec a_\mathrm{drift}$ is
\begin{equation}
\label{eq_drift}
\vec a_\mathrm{drift}=0
\end{equation}
\item
\textbf{Free-falling} The object goes into free-falling under gravity. The accelerated velocity $\vec a_\mathrm{g}$ is 
\begin{equation}
\label{apple_equation}
\vec a_\mathrm{g}=
\begin{bmatrix} 
0 \\
-g
\end{bmatrix}
\end{equation}
$g$ is the gravitational acceleration constant.
\item
\textbf{Parabola} The object moves along a parabola under gravity. The accelerated velocity is the same as $\vec a_\mathrm{g}$ defined in Eq. \ref{apple_equation}, while the object has a random initial horizontal velocity $v_x$.
\item
\textbf{Slope} 
The object slides downhill on a smooth slope. The accelerated velocity $\vec a_\mathrm{slope}$ is
\begin{equation}
\vec a_\mathrm{slope}=
\begin{bmatrix} 
g \sin \theta \cos \theta\\
-g \sin^2 \theta 
\end{bmatrix}
\end{equation}
where $\theta$ is the slope gradient.
\item
\textbf{Spring} The object is connected to a horizontal wall with a visible spring obeying Hooke's law. The accelerated velocity $\vec a_\mathrm{spring}$ is
\begin{equation}
\label{eq_spring}
\vec a_\mathrm{spring}=
\begin{bmatrix} 
0 \\
-k \cdot \left(d_y-D-X \right)/m
\end{bmatrix}
\end{equation}
The Hooke's constant $k$, attachment point y-coordinate $D$, and equilibrium distance $X$  are constants in an experiment.
\end{itemize}

For each physical scenario, we generate 300 videos for training the Observer Engine, and 100 videos for testing our pipeline, where each video has 100 frames. To simulate the real-world scenarios, by following \cite{watters2017visual} we use a random Cifar-10 \cite{krizhevsky2009learning} natural image as the background of each synthetic video. There is no overlap of background images between training set and testing set. The image size of a video frame is set as 38K$\times$38K because a larger image size enables a smaller relative error when estimating the object position. The sizes of objects in videos are the same. As for the constant independent variables, we fix $g=9.8$, $k=2$, $D=-15\,000$, and $X=5\,000$ in the experiments. We do not fix the slope gradient $\theta$ as it is an observable variable. The object initial position, the object initial velocity, object mass, and the slope gradient are random in every video.

\subsection{Learning Details}
\label{learning_detail}
In the Observer Engine, we use a two-stage Faster-RCNN object detector whose backbone network is the pretrained ResNet-101 \cite{he2016deep} model. In each stage, 2,000 and 1,000 images are randomly sampled as training set and validation set respectively. In the first stage, we train an object detector to detect objects on the original video frames (38K$\times$38K pixels). The object detector is trained by the SGD optimizer with a learning rate of 0.005, a batch size of 4, and a learning rate decay of 8. After 4 epochs of training, the detection model gets converged and obtains a 95.5\% MAP on validation set. In the second stage, we crop a 4K$\times$4K part from the original image with the bounding box output by the first stage. Then we train another object detector to refine to a more precise object position. The detector is trained by the SGD optimizer with a learning rate of 0.001, a batch size of 1, and a learning rate decay of 4. The detection model gets converged after 6 epochs and obtains a 97.7\% MAP on validation set. The object position is estimated as the center point of the bounding box output by the second detector. Table \ref{observer_baseline} shows that the euclidean distance error of our estimated position is less than 1 pixel in average.

In the Physicist Engine, we use genetic programming to evolve the syntax tree which represents a mathematical equation. The independent variables include position $d_x$, $d_y$, velocity $v_x$, $v_y$, mass $m$, and time interval $\Delta t$. The mass $m$ of object is set as known by the Physicist Engine as it could be easily estimated in the real world. In scenario of Slope, the independent variables also include $\sin \theta$ and $\cos \theta$. We do not use $\theta$ as independent variable, because the difference between $\theta$ and $\sin \theta$ is numerically trivial for regression under small $\theta$. The arithmetic operations, including addition ($+$), subtraction ($-$), multiplication ($\times$), and division ($/$), are used for every scenario. Genetic operations including crossover ($p=0.5$), subtree mutation ($p=0.15$), hoist mutation ($p=0.15$), and point mutation ($p=0.15$) are employed in evolution. 

\section{Results}

\subsection{Perceiving Mathematical Equation}

We show that our pipeline is able to perceive mathematical equations on a variety of physical scenarios with diverse visual appearances in Fig. \ref{qualitative_fig}. Five different physical scenarios are shown in the first column, where the objects are moving under corresponding dynamic equations. Our Observer Engine detects the bounding boxes (green) of objects, providing precise object positions to the Physicist Engine. At the right part of Fig. \ref{qualitative_fig}, we show the mathematical equations and syntax trees learned by the Physicist Engine, where $x$-component $\Delta d_x$ and $y$-component $\Delta d_y$ of displacement $\Delta \vec d$ are respectively shown in the second column and the third column. 

Fig. \ref{qualitative_fig} demonstrates that the Physicist Engine can learn dynamic equations of all the physical scenarios, even though the dynamic equations of some scenarios (Slope and Spring) are complex. Not only the symbolic relationships are correctly learned, the physical constants in mathematical equations are also accurately estimated by our method (e.g., the ground truth $g=4.9$ in scenarios of Free-falling, Parabola, and Slope; the ground truth $D+X=-10\,000$ in scenario of Spring). 

Please note that every equation in Fig. \ref{qualitative_fig} is generated based on the same independent variables (except particular arguments of environment) and the same arithmetic operations across all the scenarios. It reveals that our method is scalable to many other physical systems which are not included in the experiments of this work. In addition, our Observer Engine is effective in complex real-world background images with diverse visual appearances, indicating that our method is probably able to be applied to real-world videos in a future study.

\subsection{Baselines}

We have explored several variants of models for the Observer Engine and the Physicist Engine respectively, so as to quantitatively establish baselines for our newly proposed problem. 

For the Observer Engine, we study two baseline methods for a quantitative comparison with our used Two-stage Detector:
\begin{itemize}
\item
Single Detector: It is the basic Faster R-CNN \cite{ren2015faster} model, where the Region Proposal Network (RPN) is used for estimating the bound of object. The bound is used for computing the position of object.
\item
Detection + Segmentation: First, a Single Detector is used for getting an object bound. Then, a fully convolutional network (FCN) is used for segmenting the object in the pre-detected bound to localize a more accurate object position.
\item
Two-stage Detector: The method adopted by this work. Two Single Detectors are stacked to detect the object in a coarse-to-fine strategy.
\end{itemize}

\begin{table}[t]
\caption{Baselines of the Observer Engine}
\centering
\begin{tabular}{lc}
\toprule
\textbf{Method} & \textbf{MED} (pixels) \\
\midrule
Single Detector & 39.76  \\
Detection + Segmentation & 5.26 \\
Two-stage Detector    & \textbf{0.55} \\
\bottomrule
\end{tabular}
\label{observer_baseline}
\end{table}

\begin{figure}[t]
\includegraphics[width=1\linewidth]{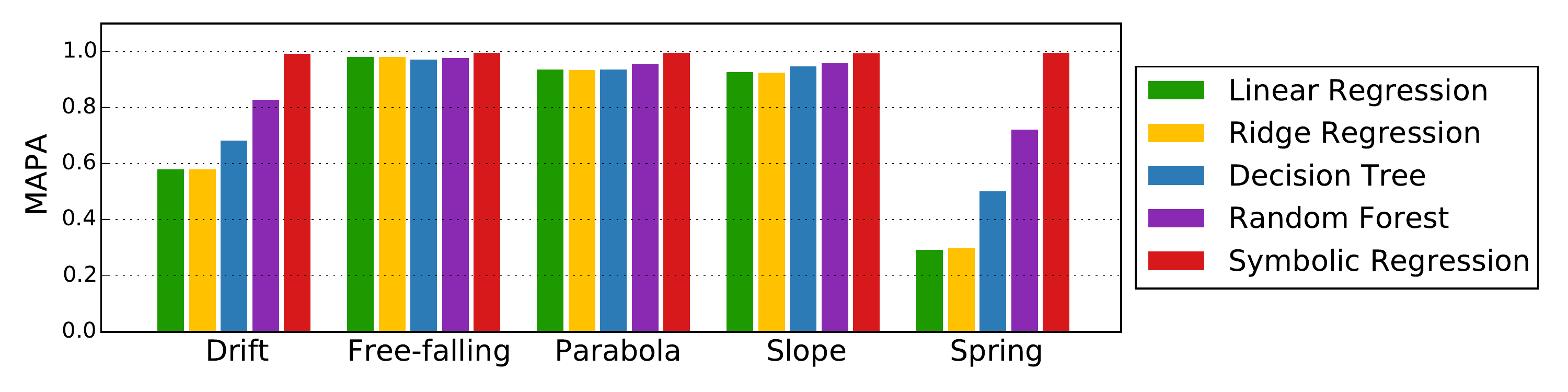}
\caption{Baselines of the Physicist Engine.
}
\label{physicist_baseline}
\end{figure}

Table \ref{observer_baseline} shows the baseline performances of the Observer Engine, under the metric of mean Euclidean distance (MED) between the estimated position and the ground-truth position. Comparing Detection + Segmentation to Single Detector, the segmentation operation is able to refine the output of single RCNN model by about 8X.  Comparing Two-stage Detector to Single Detector, the second detector successfully reduces the error by about 72X based on output of the first detector. The Two-stage Detector used in this work shows a surprising performance such that the mean error is 0.55 pixel under the 38K$\times$38K coordinate system, indicating that it can be extended to various visual scenarios and real-world applications.

For the Physicist Engine, we also study a series of common regression methods for a comparison with the symbolic regression algorithm used in this work. The baselines include (1) linear regression, (2) ridge regression, (3) decision tree, and (4) random forest. These models are implemented based on the scikit-learn toolbox \cite{scikit-learn}. Fig. \ref{physicist_baseline} shows the baseline performances of the Physicist Engine, under the metric of mean absolute percentage accuracy (MAPA) between ground-truth displacement $\Delta d$ and estimated displacement $\Delta \hat d$ as 
\begin{equation}
\text{MAPA}=1-\frac{1}{N} \sum^{N}_{i=1} \left| \frac{\Delta \hat d - \Delta d}{\Delta d} \right|
\end{equation}

MAPA denotes the relative estimation accuracy. In Fig. \ref{physicist_baseline}, baseline methods perform well on scenarios of Free-falling, Parabola, and Slope. While on scenarios of Drift and Spring, methods show distinct difference such that decision tree and random forest perform significantly better than linear regression and ridge regression. The main reason is that decision tree and random forest are non-linear models thus having much better non-linear representation capabilities than linear/ridge regression. Our symbolic regression algorithm performs the best such that its accuracy is almost 1.0 on every scene. Apparently the syntax tree of symbolic regression can perfectly represent the relationships (such as multiplication and division) between independent variables, such that symbolic regression is naturally suited for learning mathematical equations.


\begin{table}[t]
\caption{Ablation study of our pipeline (R$^2$ score). Baseline methods of the Observer Engine and the Physicist Engine are row-wise listed and column-wise listed respectively. LR: linear regression; RR: ridge regression; DT: decision tree; RF: random forest; SR: symbolic regression; GT: ground-truth equation.}
\centering
\begin{tabular}{l|ccccc|c}
\toprule
 & LR & RR & DT & RF & SR & GT \\
\midrule
Single Detector & \cellcolor[HTML]{ECF4FF}0.917 & 0.908                         & \cellcolor[HTML]{ECF4FF}0.812 & 0.932                         & \cellcolor[HTML]{ECF4FF}0.926 & 0.904                         \\

Detection + Segmentation  & 0.958                         & \cellcolor[HTML]{ECF4FF}0.958 & 0.832                         & \cellcolor[HTML]{ECF4FF}0.922 & 0.954                         & \cellcolor[HTML]{ECF4FF}0.954 \\

Two-stage Detector  & \cellcolor[HTML]{ECF4FF}0.945 & 0.944                         & \cellcolor[HTML]{ECF4FF}0.960 & 0.983                         & \cellcolor[HTML]{ECF4FF}\textbf{1.000} & 1.000                         \\ \midrule
 
Ground-truth Position & 0.945                         & \cellcolor[HTML]{ECF4FF}0.944 & 0.970                         & \cellcolor[HTML]{ECF4FF}0.984 & 1.000                         & \cellcolor[HTML]{ECF4FF}1.000\\
\bottomrule
\end{tabular}
\label{joint_baseline}
\end{table}

Table. \ref{joint_baseline} shows a more comprehensive ablation study of our pipeline, where the baseline methods of two engines are pairwise combined to be evaluated in all the physical scenarios. We use $\text{R}^2$ coefficient score as the metric to evaluate the fitting goodness in this study. It is interesting that when working with Single Detector or Detection + Segmentation, sometimes the methods of Physicist Engine perform better than the ground-truth equation. It is mainly because these methods eliminate some position errors in fitting. We observe that our pipeline (a combination of Two-stage Detector and SR) gets an 1.000 R$^2$ score, as it successfully identifies all of the dynamic equations as well as accurately estimates the constants, as shown in Fig. \ref{qualitative_fig}. Comparing Two-stage Detector with Ground-truth Position and comparing SR with GT, both methods show performances close to the ground-truth, indicating that they have good compatibilities with different methods of the other engine. 

\label{sec_baseline}

\section{Discussion}
We have introduced a new problem of deriving mathematical equations from physical scenarios, taking a step toward the goal of reasoning about universal laws from a complex environment. We have presented a pipeline including an Observer Engine and a Physicist Engine to tackle this problem for the first time. In the experiments, we have shown that our pipeline is able to perceive dynamic equations on various physical scenarios whose visual appearances are quite different. Ablation studies conducted on combinations of baselines further demonstrate the effectiveness of our pipeline. In general, our pipeline is an effective template for reasoning about the physical and dynamic systems. By combining deep learning, symbolic learning, and evolutionary algorithm, we show the potential of a hybrid machine learning system for AI reasoning. We hope this work may inspire future study on inference, induction, and conceptual understanding of general AI.

In the future, an important work is to demonstrate the proposed pipeline in real-world scenarios which may have more unknown noise than the synthetic data. It will also be important to develop techniques to handle the multi-object physical system~\cite{battaglia2016interaction,watters2017visual,wu2017learning}, in which there are interactions between objects other than the dynamics of a single object. It is a challenging and meaningful task to learn equations of a composite set of dynamic laws. In addition, our pipeline is probably able to be extended to some practical applications, e.g., helping physicists to summarize and analyse the experimental data in complex visual scenarios.


\newpage
\bibliographystyle{unsrt}
\bibliography{nips_2018}

\end{document}